\documentclass[conference]{IEEEtran}
\IEEEoverridecommandlockouts
\usepackage{cite}
\usepackage{amsmath,amssymb,amsfonts}
\usepackage{algpseudocode}
\usepackage{algorithm}
\usepackage{graphicx}
\usepackage{textcomp}
\usepackage{xcolor}
\usepackage{dblfloatfix}
\usepackage{placeins}

\def\BibTeX{{\rm B\kern-.05em{\sc i\kern-.025em b}\kern-.08em
    T\kern-.1667em\lower.7ex\hbox{E}\kern-.125emX}}

\begin{document}

\title{A Comparative Analysis on the Performance of Upper Confidence Bound Algorithms in Adaptive Deep Neural Networks
}

\author{\IEEEauthorblockN{Grigorios Papanikolaou}
\IEEEauthorblockA{\textit{Department of Electrical} \\
\textit{and Computer Engineering} \\
\textit{National Technical University of}\\
Athens, Greece \\
grigoris\_papanikolaou@mail.ntua.gr}
\and
\IEEEauthorblockN{Ioannis Kontopoulos}
\IEEEauthorblockA{\textit{Department of Electrical} \\
\textit{and Computer Engineering} \\
\textit{National Technical University of}\\
Athens, Greece \\
ikontopoulos@mail.ntua.gr}
\and
\IEEEauthorblockN{Konstantinos Tserpes}
\IEEEauthorblockA{\textit{Department of Electrical} \\
\textit{and Computer Engineering} \\
\textit{National Technical University of}\\
Athens, Greece \\
tserpes@mail.ntua.gr}
}

\maketitle

\begin{abstract}
Edge computing environments impose strict constraints on energy consumption and latency, making the deployment of deep neural networks a significant challenge. Therefore, smart and adaptive inference strategies that dynamically balance computational cost or latency with predictive accuracy are critical in edge computing scenarios. In this work, we build on Adaptive Deep Neural Networks (ADNNs) that employ the Multi-Armed Bandit (MAB) framework. Current literature leverages the first version of the Upper Confidence Bound (UCB1) strategy to dynamically select the optimal confidence threshold, enabling efficient early exits without sacrificing accuracy. However, we introduce four additional Upper Confidence Bound strategies in ADNNs, namely UCB-V, UCB-Tuned, UCB-Bayes, and UCB-BwK, and perform, for the first time, a comparative study of these strategies with respect to trade-offs between accuracy, energy consumption, and latency. The proposed UCB strategies are employed on the ResNet and MobileViT neural networks, and are evaluated on the benchmark datasets of CIFAR-10, CIFAR-10.1, and CIFAR-100. Experimental results demonstrate that all strategies achieve sub-linear cumulative regret, with UCB-Bayes converging the fastest, followed by UCB-Tuned and UCB-V. Finally, UCB-V and UCB-Tuned dominate the Pareto Frontiers of accuracy-latency and accuracy-energy trade-offs.
\end{abstract}


\section{Introduction}
\footnote{This paper has been accepted for publication in IEEE SMARTCOMP 2026}
In edge computing scenarios, trade-offs concerning accuracy, energy consumption, and latency are of the essence due to resource constraints and the need for real-time responsiveness. While significant effort has been devoted to improving training efficiency and model compactness \cite{11220458}, these approaches alone are often insufficient to meet the demanding operational requirements of edge devices. In practice, the computational cost incurred during inference can hinder overall system performance, especially in continuous or high-throughput deployment settings. Therefore, there is a growing need for smart and adaptive inference strategies that dynamically adjust computational effort to the difficulty of individual inputs. Such approaches aim to preserve predictive accuracy while reducing unnecessary computation, energy usage, and response time, thereby enabling more efficient and scalable deployment of deep neural networks at the edge.


To this end, Adaptive Deep Neural Networks (ADNNs) \cite{scardapane2024conditional} have been employed in the past aiming at achieving an optimal balance between performance and latency. These neural networks can dynamically activate or deactivate sections of their computational graph conditionally, depending on the input. A well-established technique within the ADNN literature is Dynamic Width Sparsity \cite{jiang2024mixtral}, where Neural Networks can completely skip layers during execution and may exit early during inference. Skipping layers entirely and exiting early is achieved by attaching auxiliary predictors or entire branches to a non-adaptive neural network and activating them based on certain criteria. The exit decision depends on two key components. First, a confidence measure is computed from the output logits of an intermediate classifier to quantify how reliable the current prediction is. Commonly used confidence measures in the literature include the maximum softmax probability \cite{9685469} and the entropy of the output distribution \cite{9020551}. Second, this confidence measure is compared against a thresholding mechanism that determines whether the prediction is sufficiently confident to justify an early exit. The threshold may be defined statically, using a fixed value, or dynamically, adapting to factors such as network depth, input characteristics, or computational constraints. If the confidence measure satisfies the thresholding condition, inference for the given input is terminated and the intermediate prediction is returned; otherwise, computation continues to deeper layers.


A growing body of research on early exit explores the use of the Multi-Armed Bandit (MAB)\cite{bajpai_ceebert_2024,bajpai_beyond_2025,pacheco2023adaee,9499356,ju2021dynamic,hanawal2022unsupervised,splitee} framework with contextual rewards to make the exiting decision. Each ``arm'' represents a discrete inference policy -- defined by a confidence threshold that determines the exit point -- and the bandit agent learns online the optimal policy that maximizes a reward function that depends on prediction confidence and computational or communication costs. However, most of the research focuses solely on the use of the first version of the Upper Confidence Bound (UCB1) exploration-exploitation algorithm due to its simplicity and ease of implementation, which selects arms by augmenting empirical rewards with an optimism-based uncertainty term. However, this focus neglects other UCB algorithms and their distinct exploration strategies. Identifying the differences among various groups of UCB algorithms is of utmost importance for understanding the optimal setup to employ during inference under various needs and scenarios. The reasoning behind this statement lies in the observation that there are implications related to variance awareness, cost awareness, and exit aggressiveness, along with the stability of decisions over time. Therefore, leveraging different UCB algorithms is expected to result in varying optimal and suboptimal trade-offs. Building upon this observation, the contributions of our work can be summarized as follows:

\begin{itemize}

\item We introduce in Early-Exit Deep Neural Networks (EEDNNs) the use of additional UCB variants beyond UCB1. These variants are based on different exploration and exploitation terms that may take into account an arm's reward variance or the average computational cost per threshold.

\item We present experimental comparisons across different settings between UCB variants, highlighting various trade-offs in accuracy, latency and energy consumption.

\item We empirically demonstrate that the use of the MAB framework in EEDNNs used for resource-constrained scenarios, does not always result in optimal trade-offs with respect to accuracy and latency.
\end{itemize}

The rest of the paper is structured as follows. Section \ref{literature} presents the literature in EEDNNS, while Section \ref{methodology} presents the MAB framework and the UCB variants that were incorporated in it. Finally, Section \ref{evaluation} presents the experimental evaluation, and Section \ref{conclusion} concludes the merits of this work.

\section{Related Work}
\label{literature}

EEDNNs were first introduced in 2017, based on the idea that some samples can be classified or predicted with high confidence without traversing the network’s entire computational graph during inference \cite{teerapittayanon_branchynet_2017}. Leveraging such networks allows early exiting by reducing unnecessary computation for inputs that can be classified with high confidence at shallow layers, thereby offering reduced latency, as samples are processed faster on average, which in turn lowers energy consumption. Although EEDNNs are a promising solution for adaptive inference under evolving constraints, they face notable challenges. Shallow exits often produce overconfident yet less accurate predictions, offering latency gains at the expense of predictive performance. As a result, the central challenge in EEDNNs is balancing efficiency and accuracy, a trade-off influenced by exit placement and design, training strategies, and, critically, the criteria used to trigger early exits \cite{phaseena_early-exit_2024}.

Since the introduction of BranchyNet\cite{teerapittayanon_branchynet_2017}, several training approaches beyond joint training \cite{zhang_leco_2023,zhou_bert_2020} -- which minimizes the average cross-entropy loss across exit branches -- have been proposed \cite{phaseena_early-exit_2024}. For instance, knowledge distillation has been widely employed to transfer knowledge from the main classifier to Intermediate Classifiers (ICs) \cite{phuong_distillation-based_2019,liu_fastbert_2020}. For example, self-distillation can be achieved by including Kullback-Leibler divergence as part of the training objective \cite{phuong_distillation-based_2019,liu_fastbert_2020}. Other schemes involve first training the backbone network and subsequently training the attached auxiliary predictors or exit branches, either keeping the backbone parameters frozen \cite{xin_deebert_2020, xu2023lgvit} or allowing them to be fully trainable \cite{kubaty_how_2024}. Different training strategies can lead to models with varying predictive capabilities, which in turn may affect energy consumption, inference time, and accuracy \cite{kubaty_how_2024}.

Recent research has proposed using gating networks to assess a model’s reliability based on the predictive confidence of each input sample \cite{bajpai_beyond_2025}. These networks are designed to be lightweight, so as not to introduce significant inference overhead. The gating network is shared across exits and trained implicitly: its weights are conditioned on the output logits of the main EEDNN, and its outputs are incorporated into the loss function, influencing the training outcome of the EEDNN. Consequently, the reward function can incorporate both confidence and reliability, providing information about the quality of the model’s predictions.

Relying solely on confidence measures is insufficient to prevent poor predictions. As a result, risk-control methods have emerged for EEDNNs, which operate post-hoc without assumptions on the input distribution and do not alter the offline training process. To this end, several mathematical frameworks based on conformal prediction provide guarantees for bounding the risk of incorrect predictions by selecting appropriate thresholds \cite{jazbec_fast_2024}. In addition to this approach, there is research focusing on bounding risk by leveraging the sub-linear prediction regret of algorithms used to select thresholds for each input during inference \cite{bajpai_beyond_2025} along with the use of gating networks that quantify multi-exit neural network's predictive reliability. Regret is the ideal target metric to use when evaluating UCB algorithms as it quantifies the difference between the mean optimal choice based on reward and the reward of the choice that was just made. There are two types of regret, namely i) instant regret that focuses on a single time step, and ii) cumulative regret that takes into account the rewards of the best possible actions and the actions taken by a decision-making agent over time. Moreover, it is worth noting that other work addresses \cite{meronen_fixing_2024} overconfidence by explicitly quantifying uncertainty through Bayesian approaches. Uncertainty quantification and risk control are important considerations when treating optimal threshold selection as an unsupervised learning problem, leveraging tools such as the Multi-Armed Bandit (MAB) framework.

The use of the MAB framework in EEDNNs is not new and was first introduced by Ju et al. \cite{9499356,hanawal2022unsupervised}.
The application of the MAB framework in multi-exit DNNs has been evaluated in scenarios with resource-constrained devices \cite{splitee}, demonstrating favorable accuracy-latency and accuracy-energy trade-offs, as well as improved robustness. These early works primarily focus on selecting explicitly the optimal exit based on contextual rewards. More recent approaches, such as UAT \cite{bajpai_beyond_2025} and AdaEE \cite{pacheco2023adaee}, implicitly address the optimal exit selection by concentrating on determining the optimal threshold. The reasoning behind this idea is that the optimal threshold per input will lead to the choice of optimal exit. This approach is based on the assumption that a higher threshold will probably lead to exiting at a deeper layer as the model will be more confident. To be more precise, the threshold is determined on-the-fly, rather than prior to processing inputs. UAT \cite{bajpai_beyond_2025} also introduces the theoretical foundations of bounding risk through the cumulative regret constraints of the UCB1 algorithm, in addition to utilizing gating networks to quantify predictive reliability.

All of the aforementioned frameworks essentially rely on the standard UCB1 algorithm for balancing exploration and exploitation. Consequently, aside from the contextual information incorporated in the reward function by metrics such as the model's confidence, computational cost, and communication overhead they do not account for factors such as the variance of rewards per threshold, the average cost associated with threshold selection, and the uncertainty of threshold choice. In this work, we aim to highlight the differences between UCB1 and alternative UCB algorithms that address the exploration-exploitation trade-off, illustrating their distinct strengths and weaknesses in terms of performance across various settings.

\section{Methodology}
\label{methodology}

In this section, the MAB framework and the UCB1 algorithm are presented. Moreover, alternative UCB variants are presented and incorporated into the MAB framework for the threshold selection task. Building upon previous work \cite{kaufmann_bayesian_nodate}, we introduce for the first time in the context of EEDNNs, the use of the other UCB variants.



\subsection{The UCB1 algorithm}

The UCB1 algorithm \cite{auer_finite-time_2002} addresses the exploration-exploitation trade-off by selecting at each time step or trial $t$ the arm that maximizes an Upper Confidence Bound on the expected reward as shown in equation \ref{ucb1}:

\begin{equation}
\label{ucb1}
    UCB_i(t) = \hat{\mu}_i + \sqrt{\frac{2 \log t}{N_i}}
\end{equation}

where $\mu_i$ is the empirical mean reward of arm $i$ and encourages exploitation of arms that have performed well in the past, and $N_i$ is the number of times a threshold or arm $i$ was chosen as the optimal one. On the other hand, the second term of the equation acts as an exploration bonus, assigning higher values to arms that have been sampled less frequently and decreasing logarithmically as the total number of trials $t$ increases. This formulation implements an ``optimism under uncertainty'' strategy: arms with uncertain or underexplored rewards are prioritized early on, while arms with consistently high estimated rewards are preferred asymptotically. The balance between these terms allows UCB1 to efficiently explore the action space while minimizing cumulative regret over time.

\subsection{The MAB framework}

\begin{figure}[t]
  \centering
  \includegraphics[width=\columnwidth]{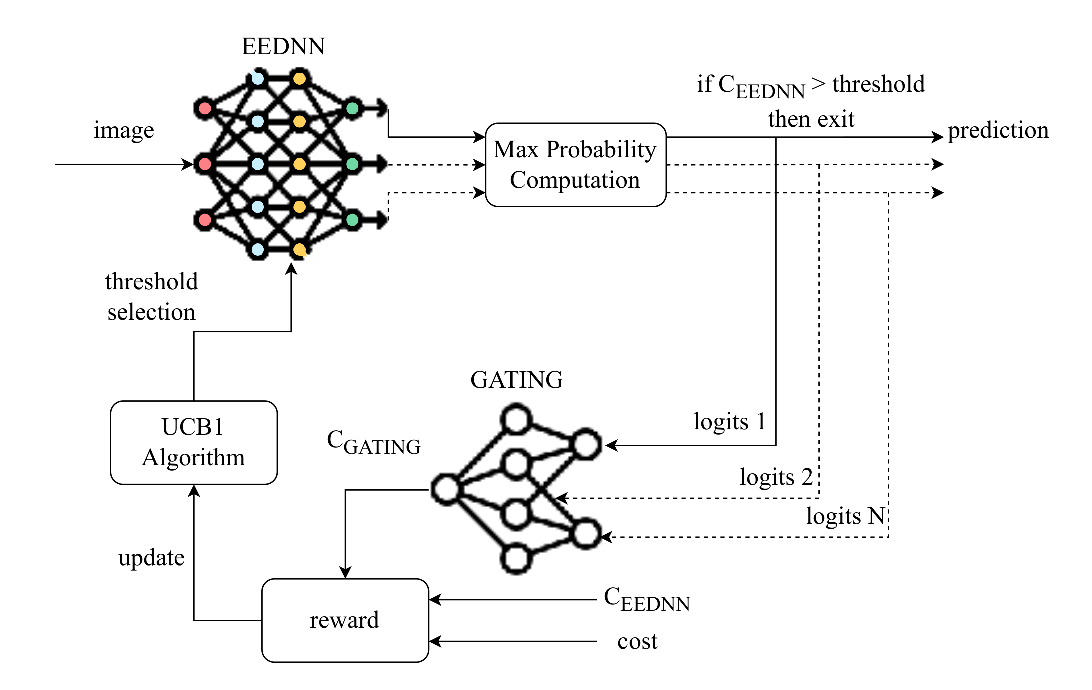}
  \caption{Unsupervised learning of optimal threshold}
  \label{fig:overview}
\end{figure}

The on-the-fly unsupervised learning procedure of the MAB framework can be visualized in Figure \ref{fig:overview}. Specifically, during the inference process, each time a sample arrives, the UCB1 algorithm initially selects a threshold or arm randomly, since all arms are set to zero at this stage. Then, the EEDNN provides its prediction based on the chosen threshold, and the logits of the EEDNN are provided to the gating network, which in turn produces a reliability score, indicating how reliable the prediction of the EEDNN is. The lower the reliability, the better the prediction of the EEDNN is. Both the reliability score $C_{Gating}$ and the EEDNN confidence value $C_{EEDNN}$ -- which is produced by the activation function of the EEDNN -- are passed into the reward function as shown in equation \ref{reward_function}:

\begin{equation}
\label{reward_function}
    reward = C_{EEDNN} * (1-C_{Gating}) - \lambda * cost
\end{equation}

where $reward \in [0,1]$, and $\lambda$ is the balancing factor used for risk control purposes as established by \cite{bajpai_beyond_2025} and is dependent on the user-defined error margin divided by the number of exit points. $cost$ is the index of the exit point chosen in the EEDNN. The higher the values of the exit point, the greater the computational cost of the EEDNN prediction is, since more layers of the EEDNN were activated. The reward value as calculated by equation \ref{reward_function} is then used to update $\hat{\mu}_i$ of equation \ref{ucb1} to calculate the Upper Confidence Bound of the selected arm. The process is repeated for each sample that is fed into the EEDNN.

\subsection{UCB-V}

The first variation of the UCB algorithm is UCB-V, which was introduced in the same work that addressed the Multi-Armed Bandit problem \cite{auer_finite-time_2002}, along with UCB1 and UCB-Tuned. UCB-V differs from UCB1 mainly because it includes the empirical variance in the main exploration term, along with an additional secondary exploration term. The exploration term of UCB-V is variance-aware, meaning that the confidence width for each arm is smaller when the observed variance $\sigma_{i}$ is low. More precisely, low-variance arms are explored less aggressively, whereas high-variance arms maintain larger confidence intervals over many pulls. This facilitates faster commitment to stable, yet still optimal, arms. Regarding the second exploration term, it is used to compensate for errors in the variance estimation, preventing the algorithm from becoming overconfident and prematurely favoring arms during the early time steps. $UCBV_i$ for the $i_{th}$ arm is computed as follows: 

\begin{equation}
\label{ucb_v}
    UCBV_i(t) = \hat{\mu}_i + \sqrt{\frac{2 \hat{\sigma}_i^2 \log t}{N_i}} + \frac{3 B \log t}{N_i}
\end{equation}

where variable $B$ is dependent on the range of rewards (i.e., the difference between the maximum and minimum possible rewards). In the context of this work $B = 1.0$ because $reward \in [0,1]$.

\subsection{UCB-Tuned}

UCB-Tuned \cite{auer_finite-time_2002} is similar to UCB-V in that both algorithms incorporate empirical variance, allowing low-variance arms to be exploited more effectively than in UCB1. However, this algorithm uses a capped variance value $V_{tilde}$ along with an additional term designed to prevent early overconfidence. As a result, if the actual variance exceeds $0.25$, the algorithm adopts a more conservative strategy, and its exiting behavior becomes less aggressive. The following equations describe how $UCBTuned_{i}$ is computed for the $i_{th}$ arm.

\begin{equation}
\label{ucb_tuned_vtilde}
             V_{\text{tilde}} \gets \min(V_i + \sqrt{2 \log t / N_i}, 0.25)
\end{equation}

\begin{equation}
\label{ucb_tuned}
            UCBTuned_i(t) = \hat{\mu}_i + \sqrt{(\log t / N_i) V_{tilde}}
\end{equation}

As discussed in the previous section, in the context of unsupervised learning for optimal threshold selection, reward components such as confidence, reliability, and computational cost are depth-dependent on average with respect to the model's layers.



\subsection{UCB-Bayes}

Building upon previous work \cite{kaufmann_bayesian_nodate}, we introduce, for the first time in the context of EEDNNs, the use of the UCB-Bayes algorithm (see Algorithm \ref{alg:bayesucb}). UCB-Bayes differs from UCB1 in that it incorporates priors while explicitly accounting for Bayesian uncertainty instead of frequentist uncertainty (e.g. UCB1). It is assumed that there are Gaussian bandits with unknown mean reward and variance of rewards. In the context of this work to efficiently approximate closed-form posteriors, of the Normal distribution, Normal-Inverse-Gamma (NIG) priors are leveraged. Algorithm \ref{alg:bayesucb} takes as input a set of user-defined arms A, NIG hyperparameters ($\mu_0, \lambda_0, \alpha_0, \beta_0$), and a quantile exponent used for tuning the growth of the confidence quantile. For all arms, statistics such as the number of pulls $N_i$, total rewards $S_i$, and squared total rewards $S^{(2)}_i$ are initialized to $0$ (line \ref{zero_init}), and each arm is pulled at least once (line \ref{count}); as is enforced for all UCB variants. At each time step $t$, a time-dependent confidence quantile is selected, aiming to enforce diminishing exploration of arms (line \ref{quantile}), and afterwards the corresponding Gaussian quantile $z_t$ is computed (line \ref{gaus_quant}). Across time steps, posterior parameters of the NIG are updated. To be more precise, posterior mean $\mu_i$ (line \ref{mean}) and precision $\lambda_i$ (line \ref{lambda}) combine empirical observations with prior information, whereas shape $\alpha_i$ (line \ref{shape}) and scale $\beta_i$ (line \ref{scale}) parameters quantify uncertainty in reward variance. Afterwards, the standard deviation of the posterior mean estimate $\sigma_i$ is computed to quantify reward noise (line \ref{sigma}). Finally, ${UCB}_i(t)$ is computed for each arm based on posterior mean reward, the Gaussian quantile and reward variance (line \ref{u_term}). Finally, at each time step parameters $N_i$, $S_i$ and $S^{(2)}_i$ are updated (line \ref{update}).

\begin{algorithm}[t]
\caption{UCB-Bayes Algorithm}
\label{alg:bayesucb}
\begin{algorithmic}[1]
\Require Arms $\mathcal{A} = \{1,\dots,K\}$, prior parameters $(\mu_0, \lambda_0, \alpha_0, \beta_0)$, quantile exponent $\alpha$ \label{init}
\State Initialize $N_i \gets 0$, $S_i \gets 0$, $S_i^{(2)} \gets 0$ for all $i \in \mathcal{A}$ \label{zero_init}
\For{$t = 1,2,\dots$}
    \State $q_t \gets 0.95$ if $t<3$ else $1 - 1/(t (\log t)^\alpha)$ \label{quantile}
    \State $z_t \gets \Phi^{-1}(q_t)$ \label{gaus_quant}
    \ForAll{$i \in \mathcal{A}$}
        \If{$N_i = 0$} \label{count}
            \State Select $i_t \gets i$ and \textbf{break}
        \Else
            \State $\lambda_i \gets \lambda_0 + N_i$ \label{lambda}
            \State $\mu_i \gets (\lambda_0 \mu_0 + S_i)/\lambda_i$ \label{mean}
            \State $\alpha_i \gets \alpha_0 + N_i/2$ \label{shape}
            \State $\beta_i \gets \beta_0 + 0.5 (S_i^{(2)} + \lambda_0 \mu_0^2 - \lambda_i \mu_i^2)$ \label{scale}
            \State $\sigma_i \gets \sqrt{\beta_i / ((\alpha_i - 1) \lambda_i)}$ \label{sigma}
            \State $\mathrm{UCB}_i(t) \gets \mu_i + z_t \sigma_i$ \label{u_term}
        \EndIf
    \EndFor
    \If{all arms evaluated}
        \State Select $i_t \gets \arg\max_i \mathrm{UCB}_i(t)$
    \EndIf
    \State Observe reward $r_t$
    \State $N_{i_t} \gets N_{i_t} + 1$, $S_{i_t} \gets S_{i_t} + r_t$, $S_{i_t}^{(2)} \gets S_{i_t}^{(2)} + r_t^2$ \label{update}
\EndFor
\end{algorithmic}
\end{algorithm}

\subsection{UCB-BwK}

Inspired by algorithms such as UCB-Ratio, UCB-BwK, and UCB1, we propose an algorithm (see Algorithm \ref{alg:ucb_ratio}) that accounts for the cost of pulling an arm during the exploration-exploitation process, in addition to its reward \cite{badanidiyuru2018bandits, badanidiyuru2014resourceful} without having a budget metric incorporated, as is the case for the original UCB-BwK. Specifically, the algorithm selects the optimal $\frac{\mu_i}{c_i}$ value, where $\mu_i$ and $c_i$ represent the mean reward and cost of the $i$-th arm, respectively. Unlike UCB1, it does not consider reward variance, and unlike the original UCB-BwK, it does not impose a decreasing budget per arm. As the number of time steps increases, arms are pulled more and $N_{i_t}$ increases, as a result the exploration term of $UCB^r_i(t)$ decreases (line \ref{exp_rew}). Simultaneously, uncertainty in cost of reward estimates diminishes as $N_{i_t}$ is increased, as is described by the computation of term $LCB^c_i(t)$ (line \ref{exp_cost}). At the end of each iteration the best trade-off is chosen (line \ref{max1}), and mean reward $\hat{\mu}^r_{i_t}$, mean cost of reward $\hat{\mu}^c_{i_t}$ along with $N_{i_t}$ are updated (lines \ref{update1} - \ref{update2}). Consequently, the algorithm chooses the threshold that offers the best trade-off between mean reward and mean cost. This algorithm can be seen as a more sophisticated version of UCB1. However, it does not incorporate variance of rewards or Bayesian uncertainty into its policy. Another limitation is that the denominator can take extremely small values; to address this, a time step-dependent term is introduced (see line \ref{t_term}) to ensure stability and is incorporated in the computations of $LCB^c_i(t)$ (line \ref{exp_cost}). The reward function of this algorithm does not incorporate the computational cost, as it already considers the cost of reward explicitly. Therefore, the confidence reliability metric provided by the gating network is the only measure provided to minimize risk.

\begin{algorithm}[t]
\caption{UCB-BwK Algorithm}
\label{alg:ucb_ratio}
\begin{algorithmic}[1]
\Require Arms (thresholds) $\mathcal{A} = \{a_1,\dots,a_K\}$
\State Initialize $N_i \gets 0$, $\hat{\mu}^r_i \gets 0$, $\hat{\mu}^c_i \gets 0$ for all $i \in \mathcal{A}$
\State Initialize total steps $t \gets 0$
\For{$t = 1,2,\dots$}
    \If{$\exists i \in \mathcal{A}$ such that $N_i = 0$}
        \State Select $i_t \gets \arg\min_i N_i$
    \Else
        \ForAll{$i \in \mathcal{A}$}
            \State $\mathrm{conf}_i(t) \gets \sqrt{\frac{2 \log t}{N_i}}$
            \State $\mathrm{UCB}^r_i(t) \gets \hat{\mu}^r_i + \mathrm{conf}_i(t)$ \label{exp_rew}
            \State $\tau_t \gets \sqrt{\frac{\log t}{t}}$ \label{t_term}
            \State $\mathrm{LCB}^c_i(t) \gets \max\!\left(\hat{\mu}^c_i - \mathrm{conf}_i(t), \tau_t \right)$ \label{exp_cost}
            \State $\mathrm{UCB\text{-}ratio}_i(t) \gets \dfrac{\mathrm{UCB}^r_i(t)}{\mathrm{LCB}^c_i(t)}$
        \EndFor
        \State Select $i_t \gets \arg\max_i \mathrm{UCB\text{-}ratio}_i(t)$ \label{max1}
    \EndIf
    \State Observe reward $r_t$ and cost $c_t$
    \State $N_{i_t} \gets N_{i_t} + 1$ \label{update1}
    \State $\hat{\mu}^r_{i_t} \gets \hat{\mu}^r_{i_t} + \dfrac{r_t - \hat{\mu}^r_{i_t}}{N_{i_t}}$
    \State $\hat{\mu}^c_{i_t} \gets \hat{\mu}^c_{i_t} + \dfrac{c_t - \hat{\mu}^c_{i_t}}{N_{i_t}}$ \label{update2}
\EndFor
\end{algorithmic}
\end{algorithm}

\section{Experiments}
\label{evaluation}




\subsection{Architectures and Datasets}

The evaluation of the proposed UCB variants was conducted across varying architectures with varying representational capacities and numbers of parameters, including Convolutional Neural Networks (CNNs) such as ResNet18, ResNet34, and ResNet50, and a CNN-Transformer architecture, namely MobileViT, which combines convolutional layers with transformer blocks. Architectures similar to MobileViT may achieve high-confidence predictions at shallower depths or lower thresholds compared to conventional CNNs as is demonstrated empirically \cite{mehta2021mobilevit}, potentially allowing earlier exits while maintaining accuracy. To sufficiently evaluate the chosen architectures with the proposed UCB variants, the following Neural Networks (NNs) were also employed to act as a baseline comparison:

\begin{itemize}
    \item ResNet18, ResNet34, ResNet50 and xxs-MobileViT default models.
    \item Early-Exit (EE) or adaptive versions of the default models (static threshold approaches).
    \item Early-Exit versions employing the MAB framework with the different UCB variants (dynamic threshold approaches).
\end{itemize}

The datasets used for evaluation purposes are CIFAR-10 \cite{krizhevsky2009learning} -- one of the most widely used datasets for image classification and is well suited for small-scale architectures -- that contains 50,000 training images and 10,000 test images; CIFAR-100 \cite{krizhevsky2009learning} containing 50,000 samples used for training and 10,000 images used for testing; and CIFAR10.1v6 \cite{recht2018cifar10.1} which extends the CIFAR-10 test set by 2,000 samples. These samples are drawn from the TinyImage dataset and are selected to induce minimal distribution shift.



\subsection{Setup}

ResNet \cite{he2016deep} and xxs-MobileViT \cite{mehta2021mobilevit} default models were trained without additional exit branches to be used as baselines. In addition, EE ResNet versions with multiple exit branches were trained using joint optimization, which is the most commonly used technique for training EE models. ResNet models were trained using the CIFAR-10 dataset. The training objective is based on the joint optimization of cross-entropy losses across all branches. As already stated, different training regimes can lead EEDNNs of the same architecture to behave differently \cite{kubaty_how_2024}. Therefore, the MobileViT model was trained in two stages. At the first stage, the default xxs-MobileViT model was trained. Then, at the second stage, exit branches were attached, and a joint optimization procedure followed. This training procedure is not only more efficient since the parameters of the exit branches are only updated, but it can also provide insights about the conclusions of this work across a more diverse landscape. Two MobileViT models were created, one trained on CIFAR-10 for the first set of experiments, and one trained on CIFAR-100 and employed at the second set of experiments that follow. The total number of exit branches is four for the ResNet models and three for MobileViT. This is because one exit branch was attached after each foundational block in each architecture (residual blocks for the ResNet architecture, and convolutional layers followed by a Transformer block for the MobileViT architecture). Finally, the aforementioned architectures were also trained in conjunction with a gating network as stated in Section \ref{methodology}. The reward function is used consistently across all experiments, and the balancing factor $\lambda$ is computed by setting the probability of incorrect classification to $\epsilon = 0.01$, relative to the number of exits as it is approximated by Bajpai et al. \cite{bajpai_beyond_2025} to control risk.


\begin{figure*}[t]
    \centering
    \includegraphics[width=\textwidth]{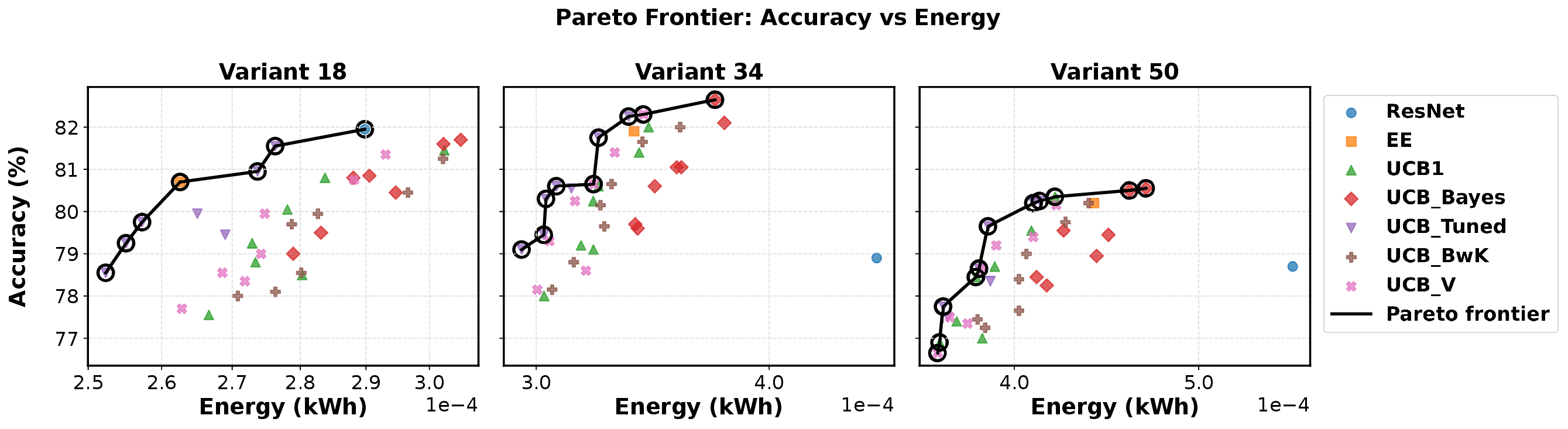}\\
    \includegraphics[width=\textwidth]{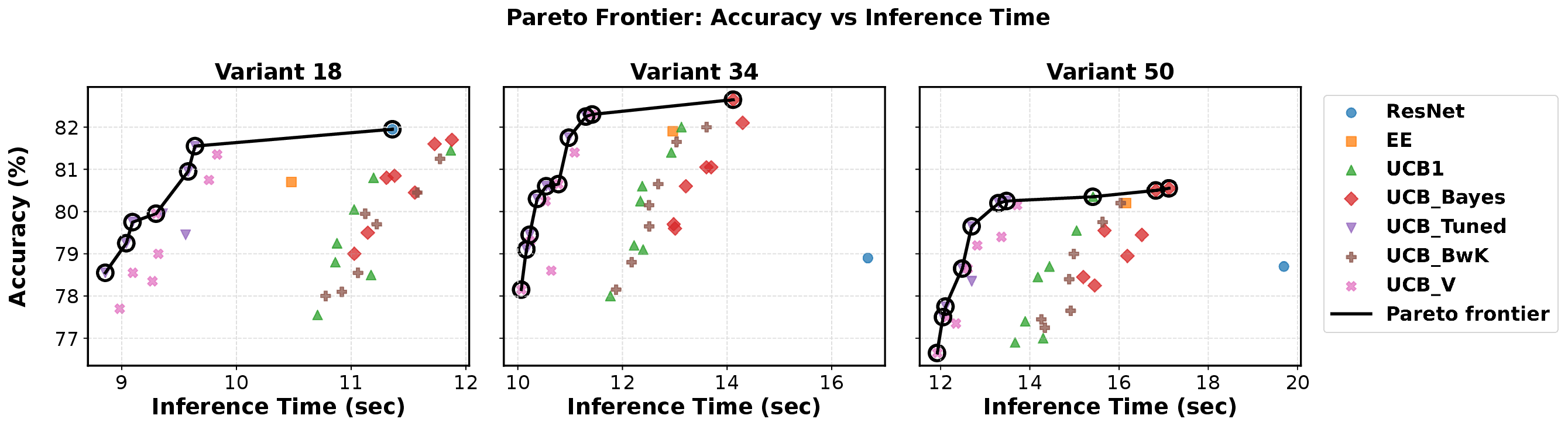}
    \caption{Performance of UCB algorithms on accuracy-energy (top) and accuracy-latency (bottom) tradeoffs across different sets of arms and ResNet variants. Inference time refers to the processing time of all samples in the CIFAR-10.1v6 test set.}
    \label{fig:resnet}
\end{figure*}

\subsection{Results}

\begin{figure}[t]
    \centering
    \includegraphics[width=\columnwidth]{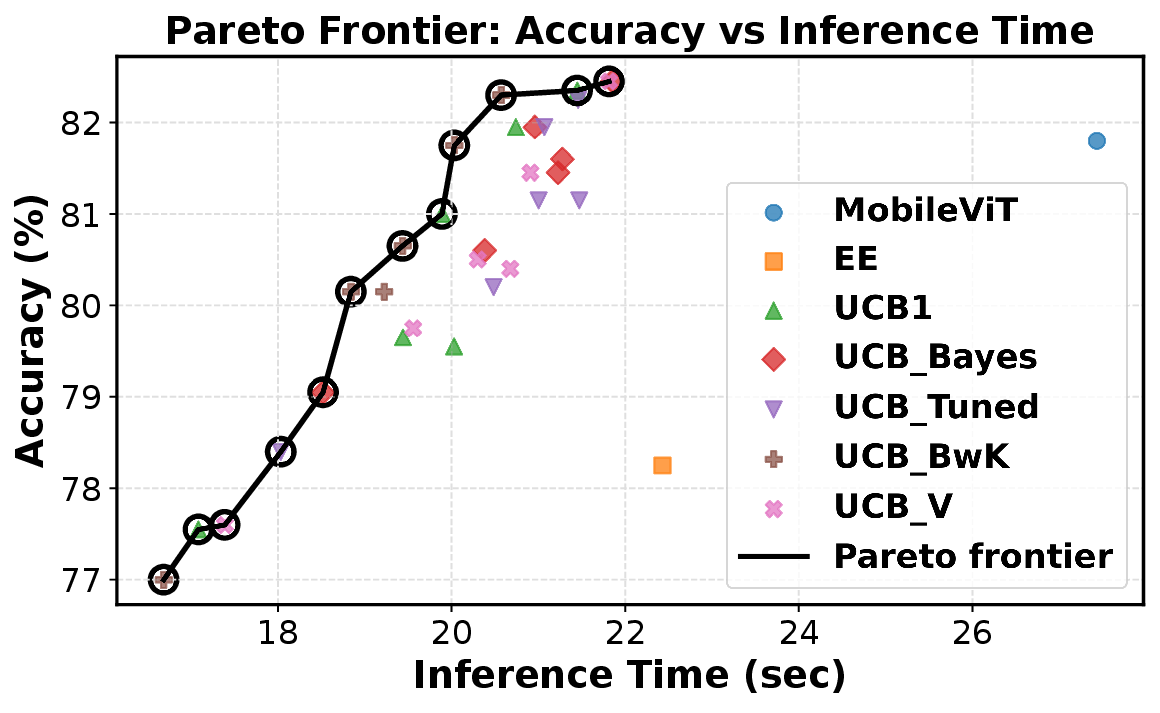}
    \caption{This figure presents the performance of UCB algorithms on the accuracy-latency tradeoff across different sets of arms leveraging MobileViT}
    \label{fig:mlatency}
\end{figure}

\begin{figure}[t]
    \centering
    \includegraphics[width=\columnwidth]{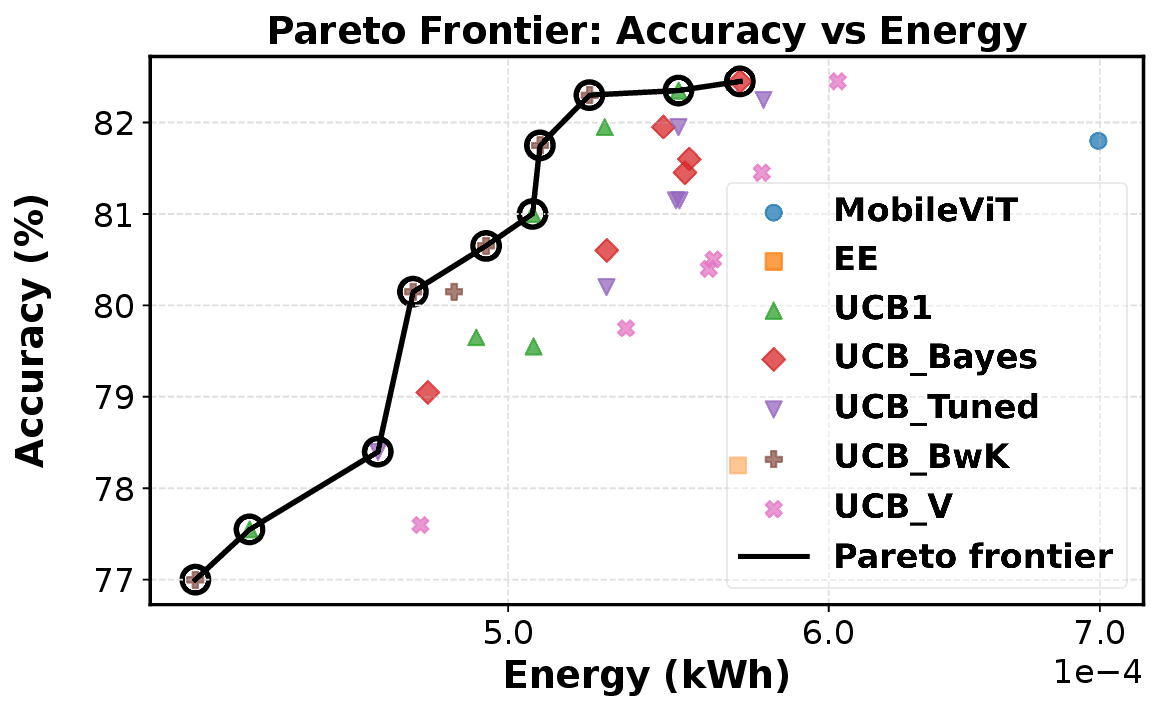}
    \caption{This figure presents the performance of UCB algorithms on the accuracy-energy tradeoff across different sets of arms leveraging MobileViT}
    \label{fig:menergy}
\end{figure}

Figure \ref{fig:resnet} visualizes the prediction accuracy against a) the energy consumed (top), and b) the inference time or latency achieved (bottom) by the ResNet variants and for each Neural Network (NN). The energy is measured in kWh and is calculated using the CodeCarbon Python library. CodeCarbon can be utilized in the scope of this work because the main purpose is to compare policies, so estimations of energy consumed are sufficient -- instead of exact energy consumption measurements --  and is treated as a proxy of computational cost. The inference time or latency is the time it takes for a model to produce an output from a given input. The illustrated line indicates the Pareto frontier and represents the solution where the accuracy cannot be improved without degrading the energy or inference time. Similarly, Figures \ref{fig:mlatency}, and \ref{fig:menergy} illustrate the trade-off of accuracy-latency and accuracy-energy, respectively, for the MobileViT NNs. For this set of experiments, the models were trained on the CIFAR-10 training set, and as a test set, the test set of the CIFAR10.1v6 dataset was employed during inference to introduce a minimal distribution shift, slightly challenging the ResNet and MobileViT ADNNs.

It can be observed from these Figures that UCB-Tuned and UCB-V dominate the accuracy-energy and accuracy-time Pareto frontiers. This is mainly because increased variance can lead to faster exploitation of low-variance arms and further exploration of high-variance arms. Therefore, fewer pulls are wasted on sub-optimal choices. Additionally, UCB-Bayes seems to result in higher accuracy at the cost of increased energy or latency. Finally, the resulting accuracy-energy and accuracy-latency Pareto frontiers indicate that CNN architectures with larger parameter counts (ResNet 34, and ResNet 50) tend to benefit more from the proposed dynamic thresholding techniques compared to static threshold approaches. Similar trends are also observed for MobileViT. However, due to MobileViT’s enhanced representational capacity, reward variance is less pronounced, which explains why the Pareto frontier is instead dominated by UCB-BwK.

To further evaluate the different versions of the NNs, another set of experiments was conducted, where the CIFAR-10 was employed for training and testing the ResNet models, and the CIFAR-100 dataset was employed for training and testing the MobileViT model. ResNet50 was chosen because of its high parameter count, whereas CIFAR-100 was included to demonstrate the extension of the conclusions drawn from the comparative analysis in datasets that are slightly more challenging. Cumulative regret across time steps was chosen for the evaluation because it provides insights on the rate of convergence to the optimal threshold to choose. The cumulative regret measures the total loss incurred by repeatedly selecting sub-optimal actions or arms instead of the optimal one. In the context of UCB-based methods, regret quantifies the cost of exploration required to identify high-reward actions.




\begin{figure}[t]
    \centering
    \includegraphics[width=\columnwidth]{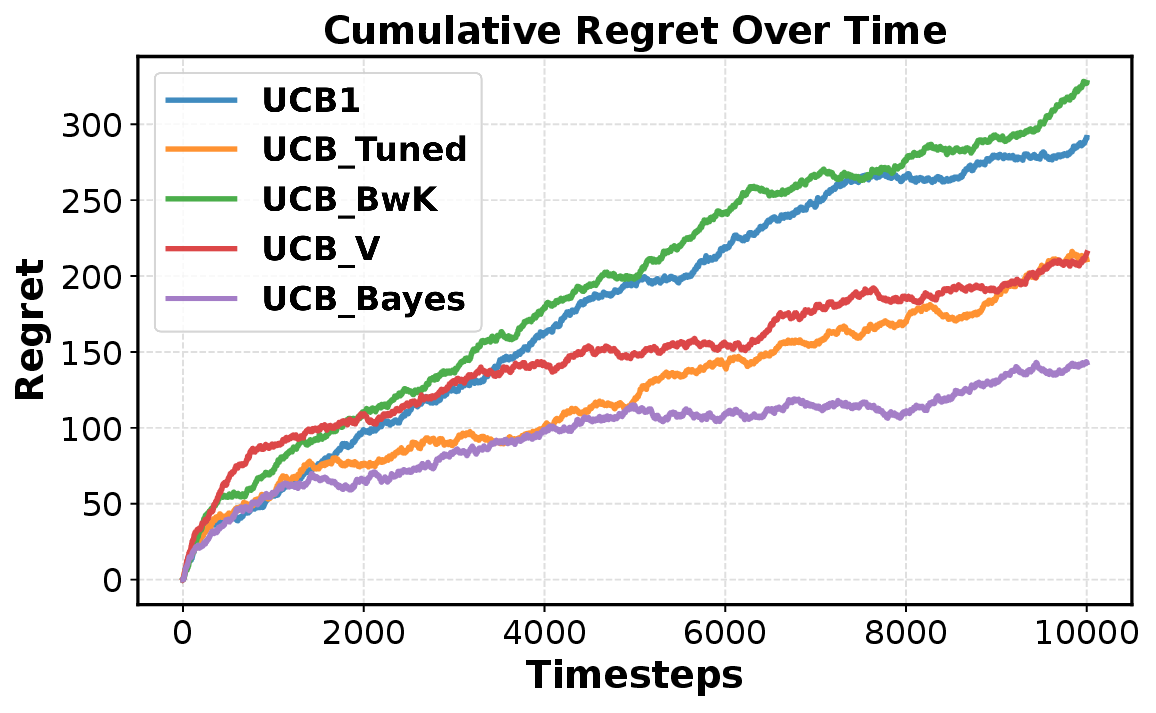}
    \caption{This figure presents the cumulative regret of UCB algorithms for  MobileViT trained on CIFAR-100}
    \label{fig:mreg}
\end{figure}

\begin{figure}[t]
    \centering
    \includegraphics[width=\columnwidth]{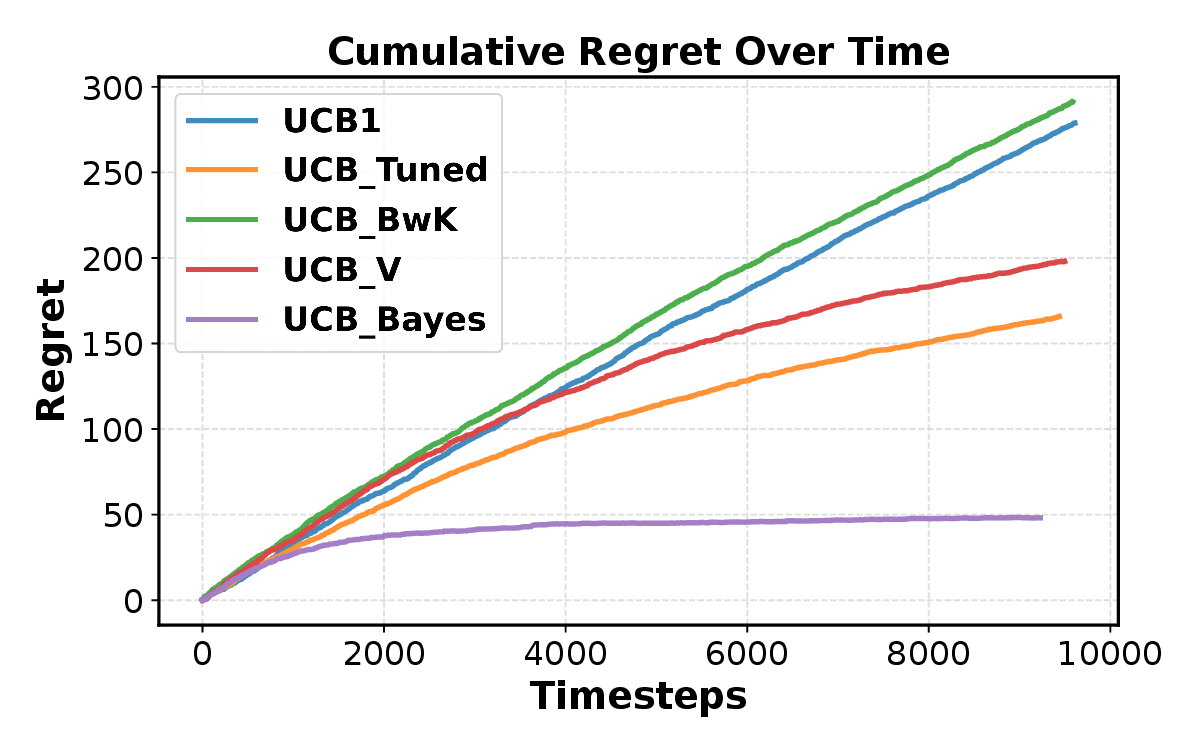}
    \caption{This figure presents the cumulative regret of UCB algorithms for ResNet50 trained on CIFAR-10}
    \label{fig:r50reg}
\end{figure}

Figures \ref{fig:mreg} and \ref{fig:r50reg} visualize the cumulative regret for MobileViT and ResNet50, respectively. For ResNet, it can be observed that the regret of UCB-Bayes does not accumulate much after approximately 4000 steps. On the other hand, the cumulative regret of UCB-V and UCB-Tuned increases at a lower rate compared to UCB1 and UCB-BwK, and does not converge as in the case of UCB-Bayes. This behavior is expected, as UCB-Bayes accounts for both the mean and variance of rewards within the framework of Bayesian uncertainty. However, the closed-form approximation of its variables introduces additional computational overhead, leading to increased latency, making it a less attractive choice for smart inference in edge intelligence against UCB-BwK, UCB-Tuned, or UCB-V. For MobileViT, similar results are observed, with the difference that the accumulation of regret is more unstable. Furthermore, the cumulative regret of UCB-Bayes does not accumulate much after approximately 6000 steps compared to the ResNet model, which does not accumulate much after 4000 steps. Finally, for both models and for all UCB variants, the regret seems to be sub-linear. For risk control guarantees to hold, regret of UCB variants should be sub-linear \cite{bajpai_beyond_2025}. UCB-Bayes, UCB1, UCB-V, and UCB-Tuned are introduced with additional guarantees of bound sub-linear regret. The above observations further support the claim that different UCB algorithms can address different requirements, highlighting the importance of selecting an appropriate variant rather than relying solely on the basic UCB1 algorithm. It is also noteworthy that UCB1 and UCB-BwK exhibit the highest average cumulative regret and converge more slowly than the other variants.

\section{Conclusions and Future work}
\label{conclusion}

In this paper we extended the MAB framework by introducing the use of multiple Upper Confidence Bound strategies for selecting the optimal threshold for early exiting over the ResNet and MobileViT architectures. Specifically, the strategies UCB-V, UCB-Tuned, UCB-Bayes, and UCB-BwK were incorporated beyond the commonly used UCB1 strategy in the literature. Furthermore, the trade-offs between these strategies were evaluated in terms of accuracy, energy consumption, and inference time. Experimental results demonstrated that UCB-Tuned and UCB-V performed the best with respect to both the energy consumption and the inference time; however, the UCB-Bayes strategy converged faster with respect to cumulative regret in both NN architectures. For future work, we intend to evaluate more UCB variants, such as LinUCB that employ the use of Neural Networks and could potentially introduce higher inference times.

\section*{Acknowledgments}
This paper has received funding from the European Union’s Horizon Europe research and innovation actions under grant agreement No 101215032. The work only reflects the authors’ views; the EU Agency is not responsible for any use that may be made of the information it contains.

\FloatBarrier
\bibliographystyle{IEEEtran}
\bibliography{refs}

\end{document}